\pdfoutput=1

\documentclass[11pt]{article}

\usepackage{acl}

\usepackage{times}
\usepackage{latexsym}

\usepackage[T1]{fontenc}

\usepackage[utf8]{inputenc}

\usepackage{microtype}

\usepackage{graphicx}
\usepackage{subfig}

\definecolor{mustard}{rgb}{1.0, 0.86, 0.35}
\definecolor{darkspringgreen}{rgb}{0.09, 0.45, 0.27}

\title{TangoBERT: Reducing Inference Cost by using Cascaded Architecture}

\author{Jonathan Mamou$^1$, Oren Pereg$^1$, Moshe Wasserblat$^1$, Roy Schwartz$^2$\\
  $^1$Intel Labs, Israel \\
  $^2$School of Computer Science and Engineering, The Hebrew University of Jerusalem, Israel \\
  $^1${\tt firstname.lastname@intel.com} \\
  $^2${\tt roy.schwartz1@mail.huji.ac.il}
}

\begin{document}
\maketitle

\begin{abstract}
The remarkable success of large transformer-based models such as BERT, RoBERTa and XLNet in many NLP tasks comes with a large increase in monetary and environmental cost due to their high computational load and energy consumption. 
In order to reduce this computational load in inference time, we present {\it TangoBERT}, a cascaded model architecture in which instances are first processed by an efficient but less accurate first tier model, and only part of those instances are additionally processed by a less efficient but more accurate second tier model.
The decision of whether to apply the second tier model is based on a confidence score produced by the first tier model. 
Our simple method has several appealing practical advantages compared to standard cascading approaches based on multi-layered transformer models. 
 
First, it enables higher speedup gains (average lower latency).
Second, it takes advantage of batch size optimization for cascading, which increases the relative inference cost reductions.  
We report TangoBERT inference CPU speedup on four text classification GLUE tasks  and on one reading comprehension task. 
Experimental results show that TangoBERT outperforms efficient early exit baseline models; on the the SST-2 task, it achieves an accuracy of 93.9\% with a CPU speedup of 8.2x.\footnote{Our code will be released upon publication.}

\end{abstract}

\section{Introduction}
The growing size of transformer-based models such as BERT \citep{BERT_devlin2019}, RoBERTa \citep{liu2019roberta} and XLNet \citep{yang2019xlnet} drives high computational load and energy consumption~\citep{strubell-etal-2019-energy,GreenAI_schwartz2019}.
In production environments, training is a one-time endeavor whereas inference occurs repeatedly for every new instance. Therefore, the inference cost is usually considerably larger than the training cost.
A recent line of work denoted `early exit'~\citep{Roy_Schwarz_early_exit_2020, wan2020orthogonalized,liu2020fastbert,xin2020deebert,xin_early_exit_2021} addresses inference cost reduction, relying on two observations; first, different model sizes incur different costs and second, instances vary in their inference `complexity'. Production instances usually range from `simple' instances for which a small and efficient model may be sufficient for accurate classification, to `complex' instances that require a deeper and larger network in order to produce accurate classification.
Early exit proposes a model that enables an enhanced speed/performance tradeoff by dynamically allowing the termination of inference at an early layer for ‘simple’ instances, while ‘complex’ instances are run through all the layers of the network. 

In this paper we present {\it TangoBERT}, a cascaded model architecture in which all the instances are processed by a small and efficient `first tier' model, and only `complex' instances are additionally processed by a larger and more accurate `second tier' model. As in early exit works \cite{Roy_Schwarz_early_exit_2020}, the confidence score produced by the first tier model determines whether or not to apply the second tier model.
The contribution of this method is threefold.
First, it enables higher inference speedup gains compared to computational load reduction approaches such as early exit; these are based on a single multi-layered  model and therefore bounded by the execution time of a single layer, which reflects the earliest possible exit. 
Second, the early exit method is often limited to using a batch size of one due to its confidence score calculation mechanism \cite{Roy_Schwarz_early_exit_2020,xin2020deebert}. In contrast, TangoBERT enables the use of larger batch sizes which further accelerates inference processing.
And third, similarly to the early exit approach but unlike other model compression approaches such as model distillation~\citep{BA_2014distillation, hinton2015distillation}, our proposed architecture provides a wide range of speed/performance tradeoff points while alleviating the need to retrain it for each point along the speed/performance curve.
We test our model on four text classification GLUE \cite{GLUE_BENCHMARK_2018} tasks (CoLA; \citealp{COLA_2018}; MRPC; \citealp{MRPC_2005}; QNLI; \citealp{SQUAD-2016}; and SST-2; \citealp{socher2013-SST2}) and on a reading comprehension task (SQuAD; \citealp{SQUAD-2016}). We show that TangoBERT outperforms DeeBERT \cite{xin2020deebert}, a popular early exit baseline model; it achieves 98\% performance of the second tier model with a CPU speedup of up to 8.2x relative to the second tier model.%

\section{Related Work}
There are two prominent approaches for inference cost reduction. The first is based on model compression, whereas the second approach is based on applying variable compute resources to inference instances based on their `complexity'. We refer to these approaches as model compression and instance-based approach, respectively.

\paragraph{Model Compression}
Model quantization \citep{Quantization_Jacob2018, Quantization_Ofir2019} is a process that compresses models by representing their parameters with fewer bits, for example, 8-bit instead of 32-bit. Another model compression method is model pruning \citep{lecun1989optimal,Pruning_NIPS2015, Pruning_ICLR2017}, a process in which some of the network’s weights are zeroed-out, producing a sparse network that reduces the computational load required for inference. Knowledge distillation \citep{BA_2014distillation, hinton2015distillation} is an efficient model compression method in which a small and efficient model is trained to mimic the behavior of a large cumbersome model. In our paper, we adopt an orthogonal approach that is instance-based. 

\paragraph{Instance-based Methods}
The second approach to inference cost reduction is based on applying variable compute resources to inference instances based on their `complexity'. 
This led to the proposal of early exit models explored by~\citet{Roy_Schwarz_early_exit_2020, wan2020orthogonalized, liu2020fastbert,xin2020deebert,xin_early_exit_2021}. %
Despite the great success of this approach, it has several limitations. First, the maximum speedup achieved by early exiting a single multi-layered transformer model is bounded by the forward pass over a single transformer layer which is the earliest exit possible.  Second, the need to make a (potentially) different early exit decision for each instance makes runtime reduction via mini-batching a challenge \cite{Roy_Schwarz_early_exit_2020}. In this paper we propose a remedy for both of these limitations.

\citet{CascadeBERT_EMNLP21} proposed cascading transformers, which can be seen as a special form of early exit, performed at the transformer level (instead of layer level). Inference runs on a series of complete transformer models with different number of layers. %
The models are executed one by one, from the smallest to the largest and the system checks, per instance, whether it can be solved by the current transformer model or needs to use a more accurate but less efficient model. 
Our work is closely related to that work; however, in their paper, the authors explored the cascading of different variations of pretrained BERT models, whereas we further investigate the cascading of transformer models in conjunction with much smaller models such as a small, distilled version of RoBERTa. %
This further reduces the inference cost and produces a wider range of points along the  performance vs.~inference speedup curve. Moreover, in this work, we explore the effects of batch size optimization on model cascading and show that using smaller first tier models increases the relative inference cost reductions.

\section{Instance `Complexity'}
\label{sec:Dataset_complexity}

Our work is motivated by the fact that instances vary in their inference `complexity'.
Following~\citet{Roy_Schwarz_early_exit_2020}, when examining different datasets, we notice that different instances pose different levels of classification `complexity'. Some instances are short, convey explicit semantics and have simple syntactic structures, while other instances are longer, include semantic subtleties and have complex syntactic structures. See, for example, the following two sentiment analysis instances from the SST-2 dataset:

\begin{enumerate} 
  \item \label{sent:simple} ``The actors are fantastic.''
  \item \label{sent:complex}``What it lacks in originality it makes up for in intelligence and B-grade stylishness.''
\end{enumerate}

A quick examination of Sentence~\ref{sent:simple} reveals that it can be correctly classified solely based on a single term (``fantastic''), whereas sentiment classification of Sentence~\ref{sent:complex} which includes contradicting terms (e.g. ``lacks in originality'' and ``B-grade'' vs.~``intelligence''), requires a deeper understanding of the syntactic structure of the sentence and of subtle semantic meanings of words and phrases for correct classification.
We refer to instances that resemble the first example as `simple' and to those resembling the second one as `complex'. We hypothesize that `simple' instances can be correctly classified using a relatively small and efficient model, whereas `complex' instances require a larger model that is able to recognize complicated syntactic patterns and subtle semantics for correct classification.

\section{TangoBERT Architecture}

\begin{figure}
    \centering
    \includegraphics[width=\linewidth]{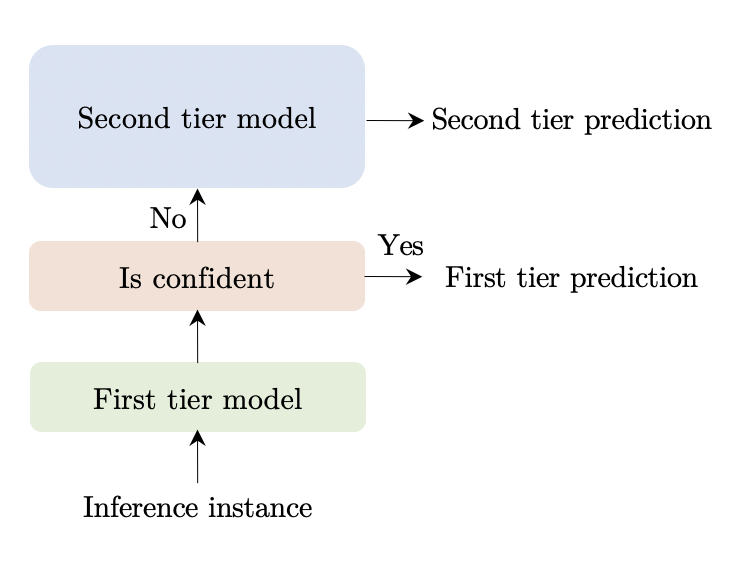}
    \captionof{figure}{TangoBERT architecture. Instance are first processed by the efficient first tier model. If this model is confident regarding its prediction, it produces the final prediction. Otherwise, the more accurate second tier model processes the instance and produces the final prediction.}
    \label{system_diagram}
\end{figure}

Our goal is to propose a simple method that offers the main advantage of the early exit approach---a wide range of speed/performance tradeoff points---while further reducing the inference computational cost as compared to that approach.
We rely on the observation that inference instances vary in their level of complexity (see Section \ref{sec:Dataset_complexity}) and this affects the amount of computational resources needed to correctly classify them.
We leverage this observation by proposing a cascaded model approach in which all inference instances are processed by an efficient but less accurate ‘first tier’ model, whereas ‘complex’ instances are additionally processed by a less efficient but more accurate ‘second tier’ model (see Figure~\ref{system_diagram}). In the case of `complex' instances, the final classification prediction is provided by the more accurate second tier model, whereas in the case of `simple' instances, the final classification prediction is provided by the more efficient but less accurate first tier model. 

The method assesses which instance is considered `complex' based on a confidence score produced by the ‘first tier’ model. Specifically, an instance will be processed also by the ‘second tier’ model if the ‘first tier’ model’s confidence score is lower than a given threshold. We use the output of the softmax function of the last hidden state of the first tier model as the confidence score. The speed/performance tradeoff point can be adjusted by changing this threshold level.

\paragraph{Discussion}

The cascaded model approach shares two appealing features with the early exit approach. First, it enables accommodating the inference computing budget of a user and provides multiple points along the speed/performance curve simply by adjusting confidence threshold level. Second, similarly to the early exit approach but unlike model distillation approaches, it does not require training a dedicated model for each speed/performance point along the curve. 
Importantly, our approach possesses two prominent advantages over the early exit approach. First, unlike the early exit approach, which is based on a single multi-layered model that bounds its maximum speedup to that of a single layer, the cascaded model approach enables much higher speedup gains due to the use of much smaller and more efficient first tier models. Second, the early exit method is limited to using a batch size of one due to its confidence score calculation mechanism \citep{Roy_Schwarz_early_exit_2020}, whereas the cascaded model approach enables the use of larger batch sizes which further accelerates inference.

\section{Experiments}
\label{sec:Experiments}

\begin{table*}	
	\centering
	\small
	
    \begin{tabular}{lccclll}%
    
    \textbf{Dataset}   & \textbf{1\textsuperscript{st}-train (K)} & \textbf{Train (K)} & \textbf{Test (K)} & \textbf{Metric} & \textbf{1\textsuperscript{st}-model}  & \textbf{2\textsuperscript{nd}-model} \\ 
    CoLA      & 172 & 8.6 &  0.5  & MCC & TinyBERT4  & BERT\\ %
    MRPC     & 768 & 3.7 & 0.5  & F1 & BiLSTM   & BERT \& RoBERTa \\ %
    QNLI      & 2095 & 104.7 & 5.5 & Accuracy & TinyBERT4   & BERT\\ %
    SQuAD    & 0 & 88.1 & 10.7 & F1 & Dynamic-TinyBERT4  & BERT\\ %
    SST-2     & 800 & 67.4 & 1.8  & Accuracy & BiLSTM   & RoBERTa     \\ %
    \end{tabular}
    \caption{Dataset statistics and TangoBERT models. %
	`1\textsuperscript{st}-train' column lists the number of unlabeled samples generated by state-of-the-art data augmentation methods and used for distilling knowledge to train the first tier model.
	`Train' column lists the number of labeled samples used for finetuning the models. 
	`Test' column lists the number of instances used for testing and comparing the different models. `Metric' column lists the metric used to evaluate performance; for CoLA, MCC metric denotes Matthew's Correlation. `1\textsuperscript{st}-model' and `2\textsuperscript{st}-model' columns list the first and and second tier models, respectively.}
	\label{table:datasets}
\end{table*}

\subsection{Datasets}
\label{sec:datasets}
We conduct experiments on various datasets with diverse state-of-the-art model performance: four text classification tasks (CoLA, MRPC, QNLI and SST-2) that are part of the GLUE benchmark~\citep{GLUE_BENCHMARK_2018}; and one reading comprehension task (SQuAD). See details below. 

\paragraph{CoLA} The Corpus of Linguistic Acceptability~\citep{COLA_2018} consists of English acceptability judgments drawn from books and journal articles on linguistic theory. Each sentence is annotated with whether it is a grammatical English sentence or not. 

\paragraph{MRPC} The Microsoft Research Paraphrase Corpus~\citep{MRPC_2005} is a collection of sentence pairs automatically extracted from online news sources. They are labeled by human annotations indicating whether or not the sentences in the pair are semantically equivalent. %

\paragraph{QNLI} This sentence pair classification task is based on the Stanford Question Answering Dataset~\citep{SQUAD-2016}, where each pair contains a question and a context sentence that may or may not contain the answer to the question. The task is to determine whether the context sentence contains the answer to the question.%

\paragraph{SST-2} The Stanford Sentiment Treebank 2~\citep{socher2013-SST2} comprises single sentences extracted from movie reviews for binary (positive/negative) sentiment classification.  

\paragraph{SQuAD} The Stanford Question Answering Dataset 1.1 \citep{SQUAD-2016} is a dataset consisting of question-paragraph pairs, where one of the sentences in the paragraph contains the answer to the corresponding question. The task is to determine the segment of text, or span, that contains the answer within the paragraph.

\paragraph{} We use the standard train-development-test splits except for datasets for which the labels of test sets are not available; in such cases, we report performance on the development set. Moreover, we use the standard metrics for each dataset. See Table~\ref{table:datasets} for dataset statistics.

\subsection{Models}
\label{sec:models}
\subsubsection{First Tier Models}
We use distilled versions of popular pre-trained models as our first-tier models:
\paragraph{BiLSTM model}~\citep{BiLSTM_Graves_2012,BiLSTM_cardie_2014} consists of pre-trained GloVe embeddings \citep{Glove_embs_2014} followed by two identical BiLSTM layers stacked one on top of the other, where the last hidden state of the second layer is followed by a Feed-Forward Network (FFN). %
\paragraph{TinyBERT model}~\citep{Tinybert_jiao_2020} is obtained by self-distilling BERT into a smaller transformer representation having fewer layers and smaller internal embedding.
\paragraph{Dynamic-TinyBERT model}~\citep{guskin2021dynamic} is a TinyBERT model that supports sequence length reduction along the layers for faster inference.

\subsubsection{Second Tier Models} We use BERT-base and RoBERTa-base as our second tier models.

\subsubsection{Early Exit Baseline Model}
We compare our TangoBERT approach to DeeBERT~\citep{xin2020deebert}, a strong early exit model. The DeeBERT implementation\footnote{\small{\url{https://github.com/castorini/DeeBERT}}} is only available for GLUE tasks, so we only use it for these tasks.
DeeBERT is applied either to BERT or to RoBERTa, depending on the compared second tier model.
Note that much like other early exit methods, DeeBERT is limited to using a batch size=1 during inference. We therefore compare it with TangoBERT with batch size=1, which is not the optimal TangoBERT setup.
We later compare the effect of using a larger batch size with TangoBERT. %

\subsubsection{Random and Oracle TangoBERT Models}

We report performance of a TangoBERT architecture that, given a speedup operating point, randomly chooses whether to process a given instance using the second tier or the first tier model. 

As an upper bound for our method, we replace the confidence score with an oracle that has access to the gold labels. %
Given a speedup operating point, the oracle chooses the combination of the first tier and second tier models that will optimize TangoBERT performance.

\subsection{Experimental Setup} 

Following~\citet{wasserblat-etal-2020-exploring}, the BiLSTM model is implemented in a fashion similar to the model described by Chollet\footnote{\small{\url{https://keras.io/examples/nlp/bidirectional\_lstm\_imdb/}}} with an embedding size of 50 and a vocabulary size of 10000. The model is fine-tuned for 3 epochs with a learning rate of \(2e^{-5}\) and a batch size of 32. The BiLSTM model is trained as a student model in a distillation setup, mimicking the behavior of RoBERTa\textsubscript{BASE}. The model contains approximately 685K parameters, meaning it is approximately 180 times more compact than RoBERTa\textsubscript{BASE}.%

For the second tier models, we adopt the HuggingFace \citep{Wolf2019HuggingFace} implementation\footnote{\small{\url{https://github.com/huggingface/transformers}}} of the BERT\textsubscript{BASE-UNCASED} and RoBERTa\textsubscript{BASE} models (both with twelve layers). 

We use TinyBERT\footnote{\small{\url{https://github.com/huawei-noah/Pretrained-Language-Model/tree/master/TinyBERT}}} and Dynamic-TinyBERT\footnote{\small{\url{https://github.com/IntelLabs/Model-Compression-Research-Package}}} models with four layers and BERT\textsubscript{BASE-UNCASED} as teacher, and we respectively denote the models TinyBERT4 and Dynamic-TinyBERT4.

We fine-tune the models for the relevant task on the standard training datasets.
For distilling knowledge for training the first tier model, we use unlabeled samples generated by state-of-the-art data augmentation methods (denoted 1\textsuperscript{st}-train); for MRPC and SST-2, we follow the data augmentation approach described by \citet{Tang_2019_DataAug} while for CoLA and QNLI, we follow the approach described by \citet{Tinybert_jiao_2020}; for SQuAD, following~\citet{guskin2021dynamic}, we did not use any data augmentation for the distillation process. 
Note that DeeBERT cannot take any advantage of the 1\textsuperscript{st}-train data, since it is unlabeled.

We run experiments on various first tier and second tier model combinations.
Note that on the whole, the first tier model is distilled from the second tier model; TinyBERT4/BERT (CoLA, QNLI), Dynamic-TinyBERT4/BERT (SQuAD) and BiLSTM/RoBERTa (MRPC, SST-2) are student/teacher cascading models. However, for MRPC, in addition to running BiLSTM/RoBERTa (where BiLSTM is distilled from RoBERTa), we run also BiLSTM/BERT (where BiLSTM is distilled from RoBERTa and not from BERT).%

To measure efficiency, we repeat each test experiment five times on CPU Intel(R) Xeon(R) CLX Platinum 8280, and report the average time.
In order to compare TangoBERT with DeeBERT, the batch size is set to one for all the inference tests reported in this section. 
TangoBERT CPU inference time is measured as the runtime when applying the first tier model to all instances, summed with the runtime when applying the second tier model only to instances with confidence level lower than the threshold.\footnote{Note that early exit methods like DeeBERT have a built-in advantage over TangoBERT, as their processing of the early layers saves some of the processing for the higher layers. Nonetheless, as our results show, the benefits of TangoBERT outweigh this factor.} %

\section{Results}
\label{sec:Res ults}

\subsection{Performance/Inference Speedup Tradeoff}
\label{sec:inf:speed}

\begin{figure*}
\centering
\subfloat{
  \includegraphics[width=0.3\textwidth]{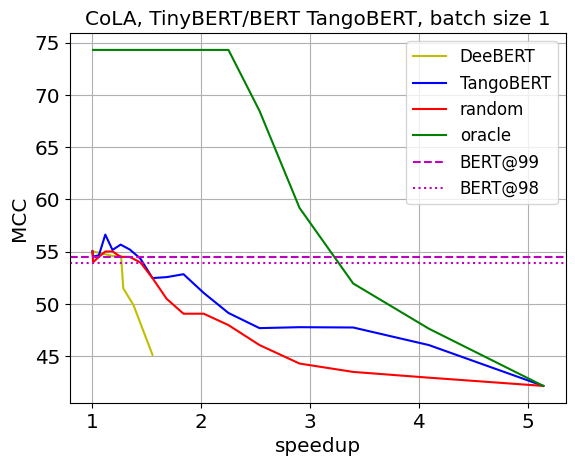}
}
\subfloat{
  \includegraphics[width=0.3\textwidth]{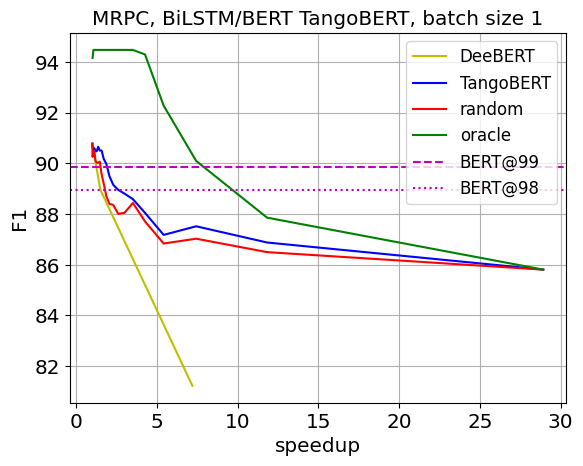}
}
\subfloat{
  \includegraphics[width=0.3\textwidth]{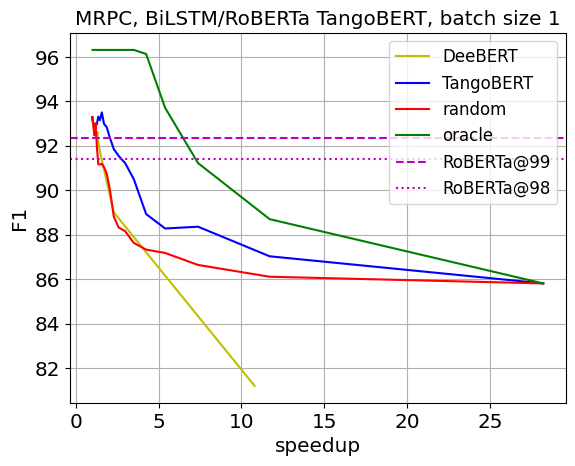}
}

\subfloat{
  \includegraphics[width=0.3\textwidth]{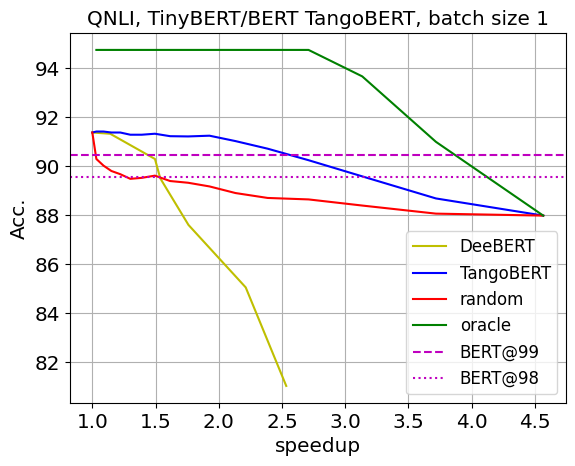}
}
\subfloat{
   \includegraphics[width=0.3\textwidth]{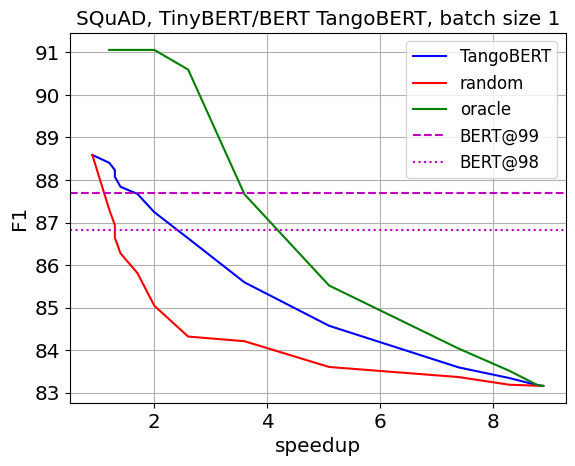}
}
\subfloat{
  \includegraphics[width=0.3\textwidth]{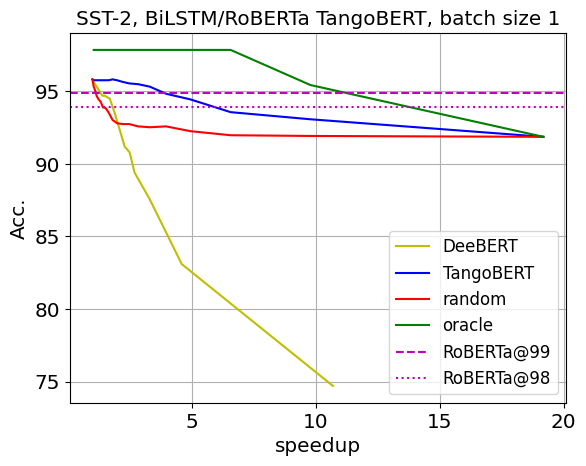}
}
\caption{Performance vs.~inference speedup achieved by the cascaded TangoBERT model (in {\textcolor{blue}{blue}}) based on the confidence score decision; DeeBERT baseline (in {\textcolor{mustard}{yellow}}); oracle TangoBERT model (in {\textcolor{darkspringgreen}{green}}) and random TangoBERT model (in {\textcolor{red}{red}}); 99\% and 98\% second tier model performance in dashed and dotted lines, respectively.}
\label{fig:exp:bs1}
\end{figure*}

Figure~\ref{fig:exp:bs1} compares the performance vs.~efficiency achieved by the cascaded TangoBERT model and our baslines.
TangoBERT and DeeBERT efficiency is measured as a function of the inference speedup in relation to the second tier model; note that the full DeeBERT model with no early exit corresponds to the second tier model. 
The leftmost point in Figure~\ref{fig:exp:bs1} represents a confidence score threshold of 1 which corresponds to using solely the second tier model (speedup 1x), whereas the rightmost point represents a confidence score threshold of 0 which corresponds to using solely the first tier model.

We first observe that, in all the different datasets tested, the TangoBERT model outperforms both the DeeBERT early exit baseline and the random TangoBERT model across all speedup points and therefore provides a better speed/performance tradeoff. %

We also report, in dashed and dotted lines respectively, 99\% and 98\% of second tier model performance (relative); {\it i.e.} 1\% and 2\% second tier model performance reduction, respectively. Table~\ref{tab:speedup:99} summarizes the speedup obtained when reaching 99\% and 98\% of the second tier performance.
We observe that TangoBERT speedup always exceeds DeeBERT speedup (for similar performance levels). At 99\%, TangoBERT achieves between 1.5x and 4.3x speedup while DeeBERT achieves between 1.2x and 1.5x speedup. At 98\%, TangoBERT achieves between 1.5x and 6.1x speedup while DeeBERT achieves between 1.3x and 1.8x speedup.  

\begin{table*}
\small
    \centering
    \begin{tabular}{c|rrrr|rrrr}
        & \multicolumn{4}{c|}{\bf speedup@99} &  \multicolumn{4}{c}{\bf speedup@98} \\
         & {\bf DB} &{\bf random} &	{\bf TangoBERT} &	{\bf oracle} & {\bf DB} &{\bf random} &	{\bf TangoBERT} &	{\bf oracle} \\ \hline
        CoLA	& 1.3x	& 1.3x	& 1.5x	& 3.2x & 1.3x	& 1.5x	& 1.5x	& 3.3x\\
        MRPC (B) &	1.2x	& 1.5x	 & 2.0x &	7.9x &	1.4x & 1.8x	& 2.7x	& 9.7x\\
        MRPC (R) & 1.2x &	1.2x &	2.1x &	6.5x & 1.5x	& 1.3x	& 2.9x	& 7.3x\\
        QNLI &	1.5x &	1.0x &	2.6x &	3.9x &	1.6x	& 1.6x	& 3.2x	& 4.1x \\
        SQuAD & N/A &	1.1x &	1.7x &	3.6x & N/A &	1.3x	& 2.5x	& 4.2x\\
        SST-2 &	1.4x &	1.2x & 4.3x & 11.5x &	1.8x	& 1.5x	& 6.1x	& 13.9x\\ 
    \end{tabular}
    \caption{Speedup at 99\% and 98\% (denoted, respectively, speedup@99 and speedup@98) of second tier model performance for DeeBERT (denoted DB), random, TangoBERT and oracle models for the different datasets. For MRPC, we report speedup for both the BERT and RoBERTa second tier models; they are denoted MRPC (B) and MRPC (R), respectively.%
    }
    \label{tab:speedup:99}
\end{table*}

\subsection{Batch Size Optimization}

\begin{figure*}
\centering
\subfloat{
  \includegraphics[width=0.3\textwidth]{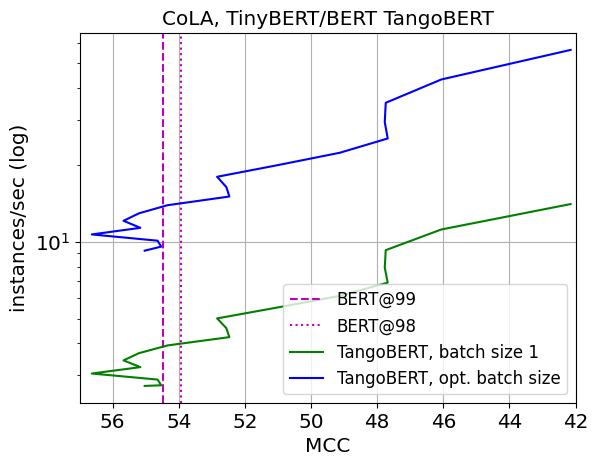}
}
\subfloat{
  \includegraphics[width=0.3\textwidth]{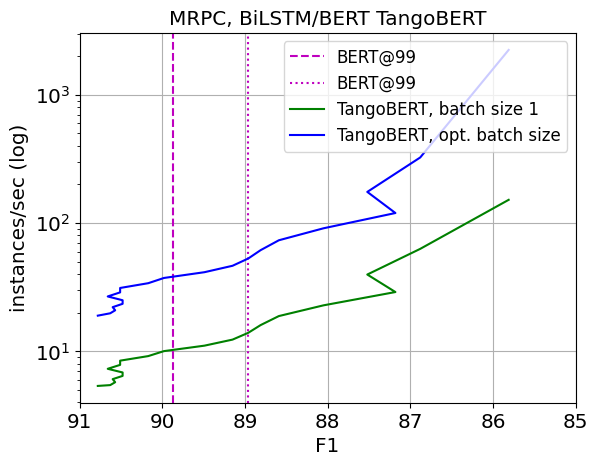}
}
\subfloat{
  \includegraphics[width=0.3\textwidth]{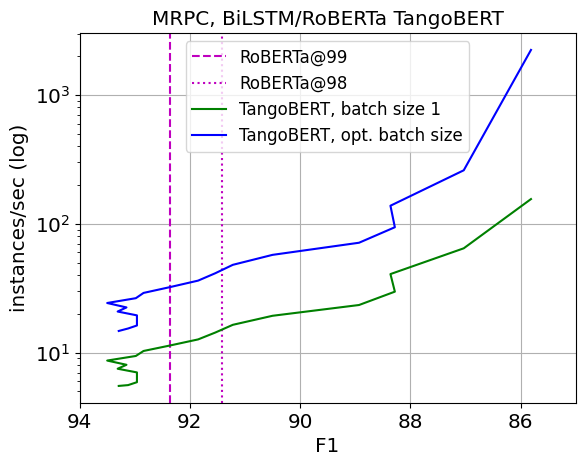}
}

\subfloat{
  \includegraphics[width=0.3\textwidth]{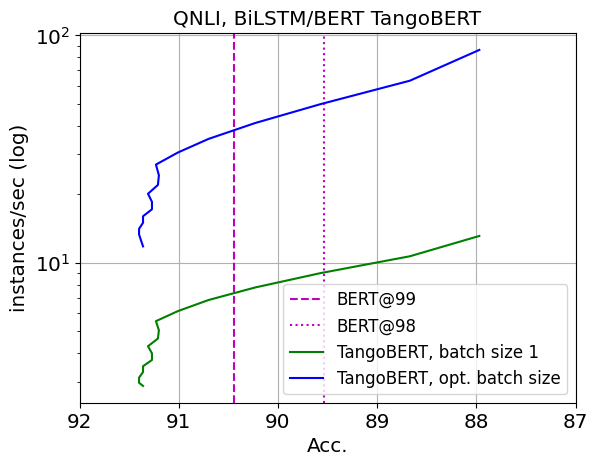}
}
\subfloat{
  \includegraphics[width=0.3\textwidth]{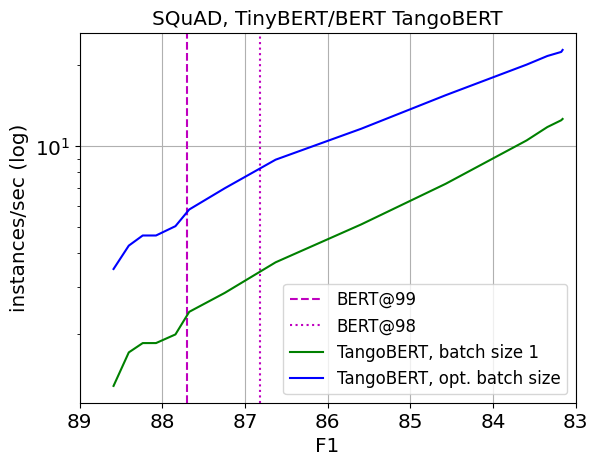}
}
\subfloat{
  \includegraphics[width=0.3\textwidth]{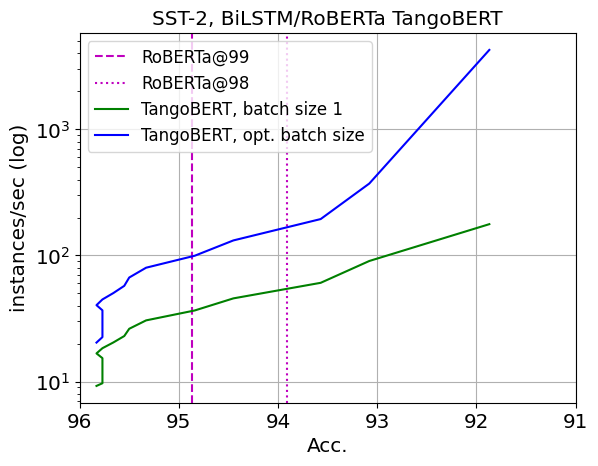}
}
\caption{Throughput (number of processed instances per second) as a function of performance for the various datasets for TangoBERT both with batch size 1 (in {\textcolor{darkspringgreen}{green}}) and with optimized batch size (in {\textcolor{blue}{blue}}). 99\% and 98\% second tier model performance are reported in dashed and dotted lines, respectively. Note that the y-axis is presented in logarithmic scale.  %
}
\label{fig:bs}
\end{figure*}

\begin{table*}
\small
    \centering
    \begin{tabular}{c|c|cc|c|cr|c|cr}
        & \multicolumn{3}{c|}{\textbf{2\textsuperscript{nd}-model}} & \multicolumn{3}{c|}{\bf TangoBERT@99} & \multicolumn{3}{c}{\bf TangoBERT@98} \\ 
        dataset & perf. & tp, bs=1 & tp, opt. bs & perf. & tp, bs=1 & tp, opt. bs & perf. & tp, bs=1 & tp, opt. bs \\\hline
        CoLA & 55.0 & 3  &  9 & 54.5	& 4	& 14 (1.6x) & 53.9 & 4 & 14 (1.6x) \\
        MRPC (B) & 90.8 & 5  &  19 & 89.9 &	10  & 38 (2.0x) & 89.0 & 14 & 52 (2.7x) \\
        MRPC (R) & 93.3 & 6  &  15 & 92.4 & 11 &	31 (2.1x) & 91.4 & 15 & 43 (2.9x) \\
        QNLI & 91.4 &  3  &  12 & 90.4 & 7 &	38 (3.2x) & 89.6 & 9 & 50 (4.2x)  \\
        SQuAD & 88.2 &  1  &  3 & 87.3 & 2 &	6 (2.0x) & 86.4 & 3 & 8 (2.7x)  \\
        SST-2 & 95.8 &  9  &  20 & 94.9 &	36 & 95 (4.8x) & 93.9 & 53 & 163 (8.2x)  \\ 
    \end{tabular}

    \caption{Throughput (number of processed instances per second) of the second tier model and throughput at 99\% and 98\% (TangoBERT@99 and TangoBERT@98, respectively) of second tier model performance, for TangoBERT, with batch size 1 (denoted `tp, bs=1') and with optimized batch size (denoted `tp, opt. bs'); model performance (denoted perf.); in parenthesis, inference time speedup of TangoBERT with optimized batch size relative to the second tier model.}
    \label{tab:t:99}
\end{table*}

One of the important hyperparameters for neural model inference is the batch size, which is the number of instances processed simultaneously. Batch size is limited by available memory and therefore, for smaller models, we can use larger batch sizes. In our case, we can use larger batch sizes for the first tier model than for the second.

Speedup results reported in Section~\ref{sec:inf:speed} were measured on batch size 1 in order to enable comparison with the early exit DeeBERT baseline that works only with batch size 1. In this section, we  further analyze the impact of the batch size on TangoBERT models.

Optimal batch size ({\it i.e.}, batch size that optimizes inference time) is determined on CPU empirically and separately for each model participating in TangoBERT.
All the test instances are processed by the first tier model with its optimal batch size. Instances that have to be processed by the second tier model are accumulated until the optimal batch size of the second tier model is reached, and then the batch is processed by the second tier model. 

Figure~\ref{fig:bs} shows the throughput as a function of performance for the various datasets for both TangoBERT with batch size 1 and TangoBERT with optimized batch size. Throughput is measured by the number of processed instances per second. 

We observe that for any performance level, the throughput increases when optimizing the batch size of the models participating in TangoBERT. Moreover, we observe a divergence in the throughput between TangoBERT, batch size 1 and TangoBERT, optimized batch size (remind that the y-axis is presented in logarithmic scale). The divergence can be explained by the fact that smaller first tier models consume less memory than larger second tier models, and therefore smaller first tier models benefit more from batch size optimization than larger second tier models.

Table~\ref{tab:t:99} summarizes the throughput of the second tier model, and the throughput at 99\% and 98\% of second tier model performance, for TangoBERT with batch size 1 and TangoBERT with optimized batch size. We also report model performance and, in parenthesis, the inference time speedup of TangoBERT with optimized batch size relative to the second tier model. We observe that at 99\% performance, TangoBERT achieves between 1.6x and 4.8x speedup, and at 98\% performance, it achieves between 1.6x and 8.2x speedup.

\begin{figure*}
\centering

\centering
\subfloat{
  \includegraphics[width=0.3\textwidth]{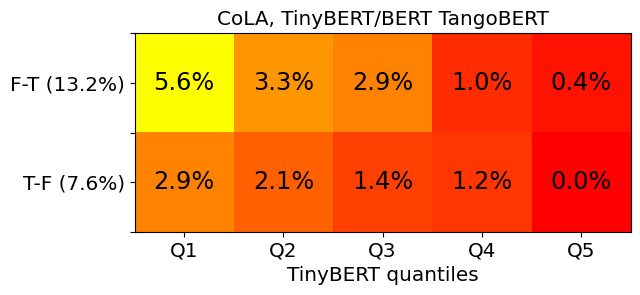}
}
\subfloat{
  \includegraphics[width=0.3\textwidth]{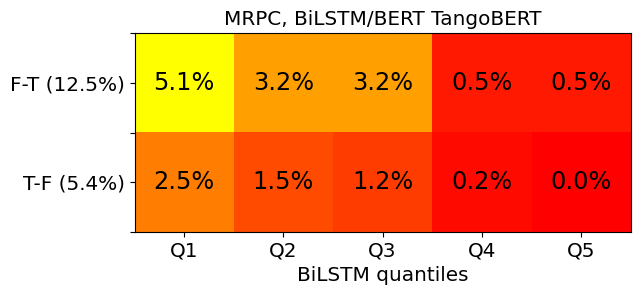}
}
\subfloat{
  \includegraphics[width=0.3\textwidth]{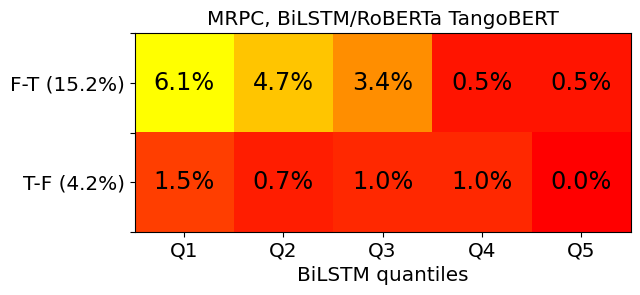}
}

\subfloat{
  \includegraphics[width=0.3\textwidth]{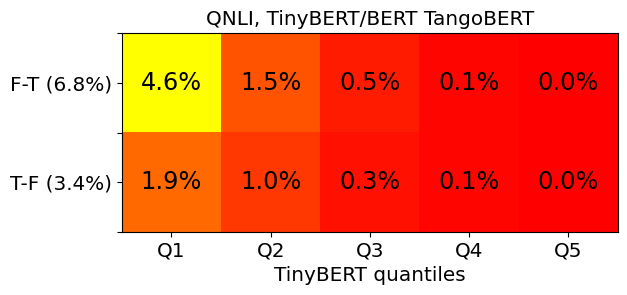}
}
\subfloat{
  \includegraphics[width=0.3\textwidth]{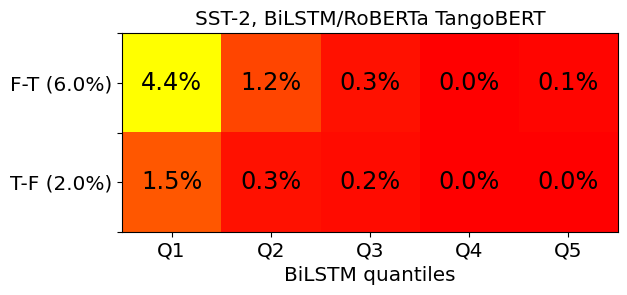}
}

\caption{Heatmaps showing for each first tier model confidence score quantile, the wrong prediction errors of the first tier model that are corrected by the second tier model (row F-T) with TangoBERT; %
the wrong predictions of the second tier model that are introduced in the correct predictions of the first tier model (row T-F); along with the percentage of the test instances that fall in each quantile (listed in the cells) and in the T-F/F-T categories (listed at the row beginning in parenthesis). Quantiles are sorted in increasing order of first tier model confidence score.}
\label{fig:err}
\end{figure*}

\section{Prediction Error Analysis}

In this section, we focus on the classification GLUE tasks and analyze the prediction errors. During TangoBERT cascading, we would ideally want prediction errors of the first tier models to be corrected by the second tier model, and correct predictions of the first tier model not be overturned by introducing wrong predictions from the second tier model.

We report in Figure~\ref{fig:err} the heatmaps, for each quantile of the test set (sorted from left to right in increasing order of the first tier model confidence score), the wrong predictions of the first tier model that are corrected by the second tier model (row F-T), and the wrong predictions of the second tier model that are introduced in the correct predictions of the first tier model (row T-F). For example, for CoLA, 5.6\% of the test instances fall in the 1\textsuperscript{st} quantile of the first tier model, are wrongly labeled by the first tier model and correctly labeled by TangoBERT; on the other hand, 2.9\% of the test instances fall in the 1\textsuperscript{st} quantile of the first tier model, are correctly labeled by the first tier model and wrongly labeled by TangoBERT. 

We observe that the F-T row appears almost in decreasing order which means that most of the first tier model prediction errors occur when the confidence score is low (on the left) and that many of them are corrected by TangoBERT during the cascading while introducing few prediction errors (T-F row). However, we also note a substantial amount of errors that are introduces by the second tier model, especially for low confidence instances. This indicates that these instances are also considerably hard for that model. Future work will study better ways to estimate the difficulty of test instances, potentially using model calibration~\cite{Degroot:1983}. 

\section{Conclusion}

In this work, we presented TangoBERT, a cascaded model approach that processes, at inference time, all instances by a small and efficient first tier model, while `complex' instances that are detected based on the first tier model’s confidence score, are additionally processed by a larger and more accurate second tier model. 
We experimented with various datasets and diverse first and second tier models participating in TangoBERT. 
We showed that TangoBERT outperforms an efficient runtime reduction baseline approach (early exit) and provides a wider range of speed/performance tradeoff point, reaching a CPU speedup of up to 8.2x with less than 2\% second tier model performance reduction. %

\bibliography{repl4nlp22}

\begin{thebibliography}{32}
\expandafter\ifx\csname natexlab\endcsname\relax\def\natexlab#1{#1}\fi

\bibitem[{Ba and Caruana(2014)}]{BA_2014distillation}
Jimmy Ba and Rich Caruana. 2014.
\newblock \href
  {http://papers.nips.cc/paper/5484-do-deep-nets-really-need-to-be-deep.pdf}
  {Do deep nets really need to be deep?}
\newblock In Z.~Ghahramani, M.~Welling, C.~Cortes, N.~D. Lawrence, and K.~Q.
  Weinberger, editors, \emph{Advances in Neural Information Processing Systems
  27}, pages 2654--2662. Curran Associates, Inc.

\bibitem[{DeGroot and Fienberg(1983)}]{Degroot:1983}
Morris~H. DeGroot and Stephen~E. Fienberg. 1983.
\newblock The comparison and evaluation of forecasters.
\newblock \emph{Journal of the Royal Statistical Society: Series D (The
  Statistician)}, 32(1-2):12--22.

\bibitem[{Devlin et~al.(2019)Devlin, Chang, Lee, and
  Toutanova}]{BERT_devlin2019}
Jacob Devlin, Ming-Wei Chang, Kenton Lee, and Kristina Toutanova. 2019.
\newblock \href {https://doi.org/10.18653/v1/N19-1423} {{BERT}: Pre-training of
  deep bidirectional transformers for language understanding}.
\newblock In \emph{Proceedings of the 2019 Conference of the North {A}merican
  Chapter of the Association for Computational Linguistics: Human Language
  Technologies, Volume 1 (Long and Short Papers)}, pages 4171--4186,
  Minneapolis, Minnesota. Association for Computational Linguistics.

\bibitem[{Dolan and Brockett(2005)}]{MRPC_2005}
William~B. Dolan and Chris Brockett. 2005.
\newblock \href {https://aclanthology.org/I05-5002} {Automatically constructing
  a corpus of sentential paraphrases}.
\newblock In \emph{Proceedings of the Third International Workshop on
  Paraphrasing ({IWP}2005)}.

\bibitem[{Graves(2012)}]{BiLSTM_Graves_2012}
Alex Graves. 2012.
\newblock \href {https://doi.org/10.1007/978-3-642-24797-2} {\emph{Supervised
  Sequence Labelling with Recurrent Neural Networks}}.
\newblock Studies in Computational Intelligence. Springer, Berlin.

\bibitem[{Guskin et~al.(2021)Guskin, Wasserblat, Ding, and
  Kim}]{guskin2021dynamic}
Shira Guskin, Moshe Wasserblat, Ke~Ding, and Gyuwan Kim. 2021.
\newblock Dynamic-tinybert: Boost tinybert's inference efficiency by dynamic
  sequence length.
\newblock \emph{arXiv preprint arXiv:2111.09645}.

\bibitem[{Han et~al.(2015)Han, Pool, Tran, and Dally}]{Pruning_NIPS2015}
Song Han, Jeff Pool, John Tran, and William Dally. 2015.
\newblock \href
  {https://proceedings.neurips.cc/paper/2015/file/ae0eb3eed39d2bcef4622b2499a05fe6-Paper.pdf}
  {Learning both weights and connections for efficient neural network}.
\newblock In \emph{Advances in Neural Information Processing Systems},
  volume~28. Curran Associates, Inc.

\bibitem[{Hinton et~al.(2015)Hinton, Vinyals, and
  Dean}]{hinton2015distillation}
Geoffrey Hinton, Oriol Vinyals, and Jeff Dean. 2015.
\newblock \href {http://arxiv.org/abs/1503.02531} {Distilling the knowledge in
  a neural network}.
\newblock Cite arxiv:1503.02531 Comment: NIPS 2014 Deep Learning Workshop.

\bibitem[{{\.I}rsoy and Cardie(2014)}]{BiLSTM_cardie_2014}
Ozan {\.I}rsoy and Claire Cardie. 2014.
\newblock \href {https://doi.org/10.3115/v1/D14-1080} {Opinion mining with deep
  recurrent neural networks}.
\newblock In \emph{Proceedings of the 2014 Conference on Empirical Methods in
  Natural Language Processing ({EMNLP})}, pages 720--728, Doha, Qatar.
  Association for Computational Linguistics.

\bibitem[{Jacob et~al.(2018)Jacob, Kligys, Chen, Zhu, Tang, Howard, Adam, and
  Kalenichenko}]{Quantization_Jacob2018}
Benoit Jacob, Skirmantas Kligys, Bo~Chen, Menglong Zhu, Matthew Tang, Andrew~G.
  Howard, Hartwig Adam, and Dmitry Kalenichenko. 2018.
\newblock Quantization and training of neural networks for efficient
  integer-arithmetic-only inference.
\newblock \emph{2018 IEEE/CVF Conference on Computer Vision and Pattern
  Recognition}, pages 2704--2713.

\bibitem[{Jiao et~al.(2020)Jiao, Yin, Shang, Jiang, Chen, Li, Wang, and
  Liu}]{Tinybert_jiao_2020}
Xiaoqi Jiao, Yichun Yin, Lifeng Shang, Xin Jiang, Xiao Chen, Linlin Li, Fang
  Wang, and Qun Liu. 2020.
\newblock \href {https://doi.org/10.18653/v1/2020.findings-emnlp.372}
  {{T}iny{BERT}: Distilling {BERT} for natural language understanding}.
\newblock In \emph{Findings of the Association for Computational Linguistics:
  EMNLP 2020}, pages 4163--4174, Online. Association for Computational
  Linguistics.

\bibitem[{LeCun et~al.(1989)LeCun, Denker, and Solla}]{lecun1989optimal}
Yann LeCun, John Denker, and Sara Solla. 1989.
\newblock Optimal brain damage.
\newblock \emph{Advances in neural information processing systems}, 2.

\bibitem[{Li et~al.(2021)Li, Lin, Chen, Ren, Li, Zhou, and
  Sun}]{CascadeBERT_EMNLP21}
Lei Li, Yankai Lin, Deli Chen, Shuhuai Ren, Peng Li, Jie Zhou, and Xu~Sun.
  2021.
\newblock \href {https://doi.org/10.18653/v1/2021.findings-emnlp.43}
  {{C}ascade{BERT}: Accelerating inference of pre-trained language models via
  calibrated complete models cascade}.
\newblock In \emph{Findings of the Association for Computational Linguistics:
  EMNLP 2021}, pages 475--486, Punta Cana, Dominican Republic. Association for
  Computational Linguistics.

\bibitem[{Liu et~al.(2020)Liu, Zhou, Zhao, Wang, Deng, and
  Ju}]{liu2020fastbert}
Weijie Liu, Peng Zhou, Zhe Zhao, Zhiruo Wang, Haotang Deng, and Qi~Ju. 2020.
\newblock \href {http://arxiv.org/abs/2004.02178} {Fastbert: a self-distilling
  bert with adaptive inference time}.

\bibitem[{Liu et~al.(2019)Liu, Ott, Goyal, Du, Joshi, Chen, Levy, Lewis,
  Zettlemoyer, and Stoyanov}]{liu2019roberta}
Yinhan Liu, Myle Ott, Naman Goyal, Jingfei Du, Mandar Joshi, Danqi Chen, Omer
  Levy, Mike Lewis, Luke Zettlemoyer, and Veselin Stoyanov. 2019.
\newblock \href {http://arxiv.org/abs/1907.11692} {Roberta: A robustly
  optimized bert pretraining approach}.
\newblock Cite arxiv:1907.11692.

\bibitem[{Pennington et~al.(2014)Pennington, Socher, and
  Manning}]{Glove_embs_2014}
Jeffrey Pennington, Richard Socher, and Christopher~D Manning. 2014.
\newblock Glove: Global vectors for word representation.
\newblock In \emph{EMNLP}, volume~14, pages 1532--1543.

\bibitem[{Rajpurkar et~al.(2016)Rajpurkar, Zhang, Lopyrev, and
  Liang}]{SQUAD-2016}
Pranav Rajpurkar, Jian Zhang, Konstantin Lopyrev, and Percy Liang. 2016.
\newblock \href {https://doi.org/10.18653/v1/D16-1264} {{SQ}u{AD}: 100,000+
  questions for machine comprehension of text}.
\newblock In \emph{Proceedings of the 2016 Conference on Empirical Methods in
  Natural Language Processing}, pages 2383--2392, Austin, Texas. Association
  for Computational Linguistics.

\bibitem[{Schwartz et~al.(2020{\natexlab{a}})Schwartz, Dodge, Smith, and
  Etzioni}]{GreenAI_schwartz2019}
Roy Schwartz, Jesse Dodge, Noah~A. Smith, and Oren Etzioni. 2020{\natexlab{a}}.
\newblock \href {https://doi.org/10.1145/3381831} {Green {AI}}.
\newblock \emph{Communications of the ACM (CACM)}, 63(12):54--63.

\bibitem[{Schwartz et~al.(2020{\natexlab{b}})Schwartz, Stanovsky, Swayamdipta,
  Dodge, and Smith}]{Roy_Schwarz_early_exit_2020}
Roy Schwartz, Gabriel Stanovsky, Swabha Swayamdipta, Jesse Dodge, and Noah~A.
  Smith. 2020{\natexlab{b}}.
\newblock \href {https://doi.org/10.18653/v1/2020.acl-main.593} {The right tool
  for the job: Matching model and instance complexities}.
\newblock In \emph{Proceedings of the 58th Annual Meeting of the Association
  for Computational Linguistics}, pages 6640--6651, Online. Association for
  Computational Linguistics.

\bibitem[{Socher et~al.(2013)Socher, Perelygin, Wu, Chuang, Manning, Ng, and
  Potts}]{socher2013-SST2}
Richard Socher, Alex Perelygin, Jean Wu, Jason Chuang, Christopher~D. Manning,
  Andrew Ng, and Christopher Potts. 2013.
\newblock \href {https://www.aclweb.org/anthology/D13-1170} {Recursive deep
  models for semantic compositionality over a sentiment treebank}.
\newblock In \emph{Proceedings of the 2013 Conference on Empirical Methods in
  Natural Language Processing}, pages 1631--1642, Seattle, Washington, USA.
  Association for Computational Linguistics.

\bibitem[{Strubell et~al.(2019)Strubell, Ganesh, and
  McCallum}]{strubell-etal-2019-energy}
Emma Strubell, Ananya Ganesh, and Andrew McCallum. 2019.
\newblock \href {https://doi.org/10.18653/v1/P19-1355} {Energy and policy
  considerations for deep learning in {NLP}}.
\newblock In \emph{Proceedings of the 57th Annual Meeting of the Association
  for Computational Linguistics}, pages 3645--3650, Florence, Italy.
  Association for Computational Linguistics.

\bibitem[{Tang et~al.(2019)Tang, Lu, and Lin}]{Tang_2019_DataAug}
Raphael Tang, Yao Lu, and Jimmy Lin. 2019.
\newblock \href {https://doi.org/10.18653/v1/D19-6122} {Natural language
  generation for effective knowledge distillation}.
\newblock In \emph{Proceedings of the 2nd Workshop on Deep Learning Approaches
  for Low-Resource NLP (DeepLo 2019)}, pages 202--208, Hong Kong, China.
  Association for Computational Linguistics.

\bibitem[{Wan et~al.(2020)Wan, Hoffmann, Lu, and Maire}]{wan2020orthogonalized}
Chengcheng Wan, Henry Hoffmann, Shan Lu, and Michael Maire. 2020.
\newblock \href {http://arxiv.org/abs/2008.06635} {Orthogonalized sgd and
  nested architectures for anytime neural networks}.

\bibitem[{Wang et~al.(2018)Wang, Singh, Michael, Hill, Levy, and
  Bowman}]{GLUE_BENCHMARK_2018}
Alex Wang, Amanpreet Singh, Julian Michael, Felix Hill, Omer Levy, and Samuel
  Bowman. 2018.
\newblock \href {https://doi.org/10.18653/v1/W18-5446} {{GLUE}: A multi-task
  benchmark and analysis platform for natural language understanding}.
\newblock In \emph{Proceedings of the 2018 {EMNLP} Workshop {B}lackbox{NLP}:
  Analyzing and Interpreting Neural Networks for {NLP}}, pages 353--355,
  Brussels, Belgium. Association for Computational Linguistics.

\bibitem[{Warstadt et~al.(2018)Warstadt, Singh, and Bowman}]{COLA_2018}
Alex Warstadt, Amanpreet Singh, and Samuel~R. Bowman. 2018.
\newblock \href {http://arxiv.org/abs/1805.12471} {Neural network acceptability
  judgments}.
\newblock \emph{CoRR}, abs/1805.12471.

\bibitem[{Wasserblat et~al.(2020)Wasserblat, Pereg, and
  Izsak}]{wasserblat-etal-2020-exploring}
Moshe Wasserblat, Oren Pereg, and Peter Izsak. 2020.
\newblock \href {https://doi.org/10.18653/v1/2020.sustainlp-1.5} {Exploring the
  boundaries of low-resource {BERT} distillation}.
\newblock In \emph{Proceedings of SustaiNLP: Workshop on Simple and Efficient
  Natural Language Processing}, pages 35--40, Online. Association for
  Computational Linguistics.

\bibitem[{Wolf et~al.(2019)Wolf, Debut, Sanh, Chaumond, Delangue, Moi, Cistac,
  Rault, Louf, Funtowicz, and Brew}]{Wolf2019HuggingFace}
Thomas Wolf, Lysandre Debut, Victor Sanh, Julien Chaumond, Clement Delangue,
  Anthony Moi, Pierric Cistac, Tim Rault, R'emi Louf, Morgan Funtowicz, and
  Jamie Brew. 2019.
\newblock Huggingface's transformers: State-of-the-art natural language
  processing.
\newblock \emph{ArXiv}, abs/1910.03771.

\bibitem[{Xin et~al.(2020)Xin, Tang, Lee, Yu, and Lin}]{xin2020deebert}
Ji~Xin, Raphael Tang, Jaejun Lee, Yaoliang Yu, and Jimmy Lin. 2020.
\newblock \href {http://arxiv.org/abs/2004.12993} {Deebert: Dynamic early
  exiting for accelerating bert inference}.

\bibitem[{Xin et~al.(2021)Xin, Tang, Yu, and Lin}]{xin_early_exit_2021}
Ji~Xin, Raphael Tang, Yaoliang Yu, and Jimmy Lin. 2021.
\newblock \href {https://www.aclweb.org/anthology/2021.eacl-main.8}
  {{BER}xi{T}: Early exiting for {BERT} with better fine-tuning and extension
  to regression}.
\newblock In \emph{Proceedings of the 16th Conference of the European Chapter
  of the Association for Computational Linguistics: Main Volume}, pages
  91--104, Online. Association for Computational Linguistics.

\bibitem[{Yang et~al.(2019)Yang, Dai, Yang, Carbonell, Salakhutdinov, and
  Le}]{yang2019xlnet}
Zhilin Yang, Zihang Dai, Yiming Yang, Jaime Carbonell, Ruslan Salakhutdinov,
  and Quoc~V. Le. 2019.
\newblock \href {http://arxiv.org/abs/1906.08237} {Xlnet: Generalized
  autoregressive pretraining for language understanding}.
\newblock Cite arxiv:1906.08237Comment: Pretrained models and code are
  available at https://github.com/zihangdai/xlnet.

\bibitem[{Zafrir et~al.(2019)Zafrir, Boudoukh, Izsak, and
  Wasserblat}]{Quantization_Ofir2019}
Ofir Zafrir, Guy Boudoukh, Peter Izsak, and Moshe Wasserblat. 2019.
\newblock Q8bert: Quantized 8bit bert.
\newblock \emph{2019 Fifth Workshop on Energy Efficient Machine Learning and
  Cognitive Computing - NeurIPS Edition (EMC2-NIPS)}, pages 36--39.

\bibitem[{Zhu and Gupta(2017)}]{Pruning_ICLR2017}
Michael Zhu and Suyog Gupta. 2017.
\newblock \href
  {http://dblp.uni-trier.de/db/journals/corr/corr1710.html#abs-1710-01878} {To
  prune, or not to prune: exploring the efficacy of pruning for model
  compression.}
\newblock \emph{CoRR}, abs/1710.01878.

\end{thebibliography}
\bibliographystyle{acl_natbib}

\end{document}